\documentclass{article}
\usepackage[skins]{tcolorbox}  
\usepackage{graphicx}  
\usepackage{xcolor}    
\usepackage{amsmath}   
\usepackage{amssymb}   
\usepackage{hyperref}
\usepackage{multirow}
\usepackage{float}
\usepackage{hyperref}
\usepackage{caption}
\usepackage{booktabs}

\usepackage{caption} 
\usepackage[labelfont=bf]{caption}

\tcbset{
    colback=green!10!gray!5,   
    colframe=green!50!gray!50, 
    arc=3mm,                   
    width=0.9\textwidth,       
    boxrule=0pt,               
    boxsep=10pt,               
    top=20pt, bottom=5pt,    
    enhanced,                  
    center                    
}

 \usepackage[a4paper, top=2cm, bottom=2.5cm, left=2.5cm, right=2.5cm]{geometry}
 
\begin{document}

\vspace*{-0.5cm}  

\begin{center} 
    \begin{tcolorbox}
    
        \vspace{-0.5cm}

 \vspace{0.3cm} 
 
          \begin{center}
    \textbf{\large F3-Net: \underline Foundation Model for \underline Full Abnormality Segmentation of Medical Images with \underline Flexible Input Modality Requirement}
\end{center}

        \vspace{1cm}  
        \textbf{Seyedeh Sahar Taheri Otaghsara} DDS,   
        \textbf{Reza Rahmanzadeh}  MD, PhD
        
 \vspace{0.4cm} 
        \raggedright {\small \textcolor{blue}{\texttt{ AI Lab, UltraAI}}} 

        \vspace{0.5cm} 

   F3-Net is a foundation model designed to overcome persistent challenges in clinical medical image segmentation, including reliance on complete multimodal inputs, limited generalizability, and narrow task specificity. Through flexible synthetic modality training, F3-Net maintains robust performance even in the presence of missing MRI sequences—leveraging a zero-image strategy to substitute absent modalities without relying on explicit synthesis networks, thereby enhancing real-world applicability. Its unified architecture supports multi-pathology segmentation across glioma, metastasis, stroke, and white matter lesions without retraining, outperforming CNN-based and transformer-based models that typically require disease-specific fine-tuning. Evaluated on diverse datasets such as BraTS 2021, BraTS 2024, and ISLES 2022, F3-Net demonstrates strong resilience to domain shifts and clinical heterogeneity. On the whole pathology dataset, F3-Net achieves average Dice Similarity Coefficients (DSCs) of 94.31\% for BraTS-GLI 2024, 82.07\% for BraTS-MET 2024, 94.12\% for BraTS 2021, and 79.92\% for ISLES 2022. This positions it as a versatile, scalable solution bridging the gap between deep learning research and practical clinical deployment.

\vspace{0.5cm} 
\raggedright {\small June 2025}  

 \vspace{0.2cm} 
        \raggedright {\small Correspondence: \textcolor{blue}{\texttt{reza@theultra.ai}}} 

   \vspace{0.3cm} 
\hspace{0.3cm} 
\hfill 
\raisebox{-0.1cm}{\includegraphics[width=2cm]{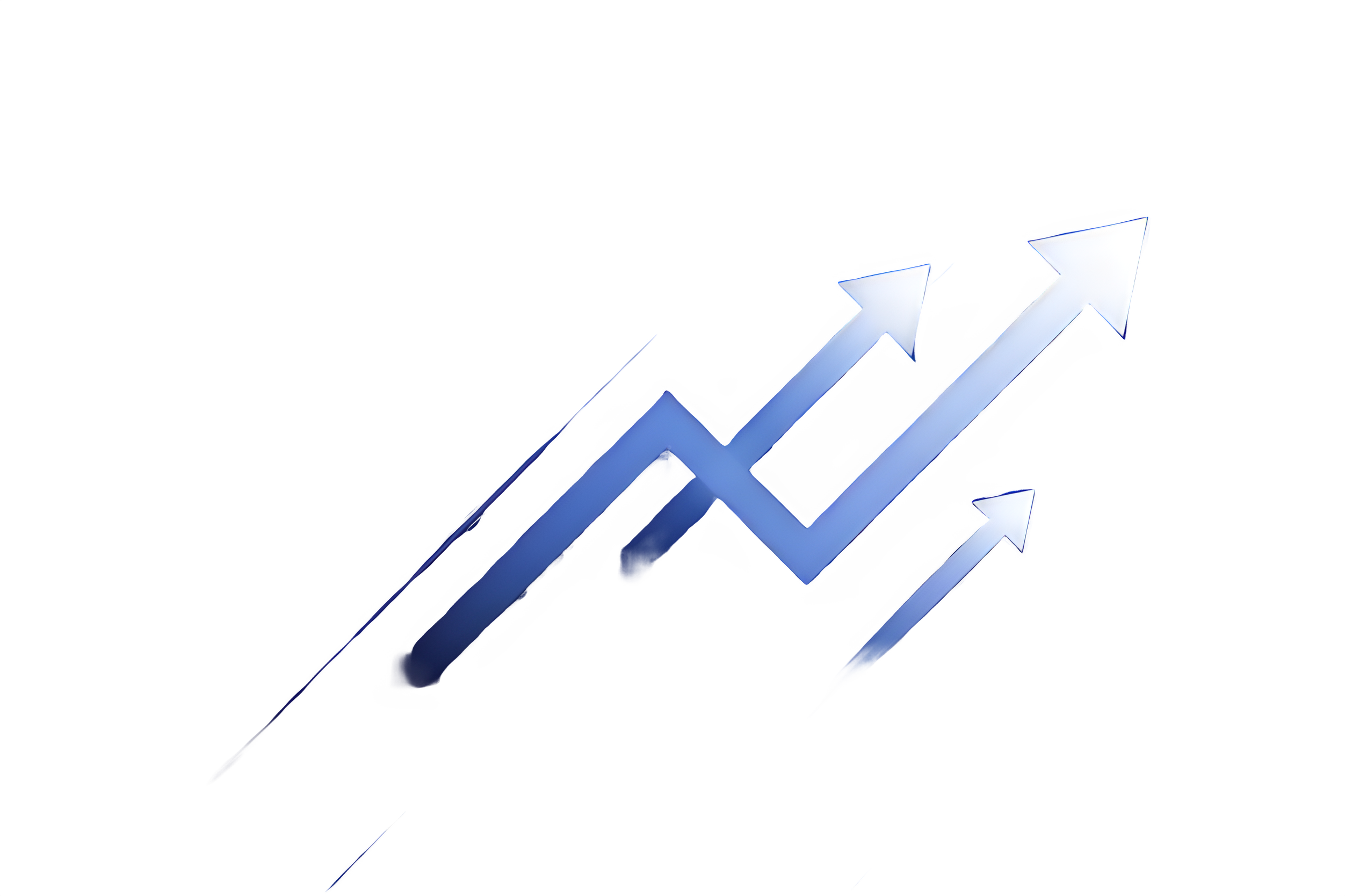}}  
\hspace{-0.5cm}  
\includegraphics[width=3cm]{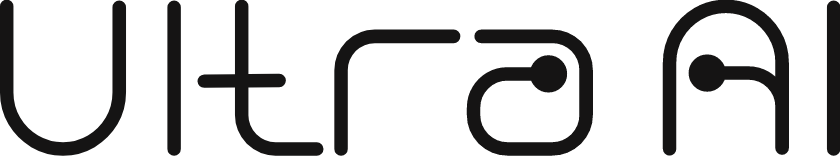}  
       
    \end{tcolorbox}
\end{center}

\vspace{0.2cm} 

\section{Introduction}

Accurate segmentation of brain abnormalities in medical imaging is crucial for diagnosis, treatment planning, and monitoring across various neurological disorders, and in this context, segmentation accuracy directly influences surgical strategies, radiotherapy planning, and disease progression assessment \cite{kiryu2023clinical, chen2023deep}.  
Deep learning-based automatic segmentation methods have shown great promise in enhancing consistency, reducing manual workload, and enabling high-throughput analysis \cite{koshino2021narrative}. However, several practical challenges still limit their generalizability and real-world applicability. In this regard, the performance of deep learning models can degrade significantly when applied to images obtained from different clinical settings, which often have variations in scanner hardware, acquisition protocols, and reconstruction algorithms \cite{kiryu2023clinical, lee2024foundation, zhang2023challenges}. These issues require sophisticated and adaptable models that can maintain robustness in diverse imaging scenarios \cite{otaghsara2025multi,zhu2024medicalsam2segment,bao2024foundation}.

One significant limitation of current segmentation models is their reliance on complete multimodal MRI inputs. However, in clinical practice, certain modalities may be missing due to patient-specific factors, scanner limitations, or time constraints, potentially reducing the effectiveness of these models. Some generative models, such as Generative Adversarial Networks (GANs), have been proposed to synthesize missing modalities, but these models often cannot recover modality-specific biological information crucial for pathology detection \cite{yang2021gan,skandarani2021gans,shomalzadeh2023generative,radiographics2020primer,liu2024vision}.  
Another significant challenge lies in the narrow focus of existing models, which are typically trained to segment a specific condition (e.g., glioma or stroke) while often failing to generalize to other, unrelated conditions. This limited focus severely restricts their clinical applicability, especially in patients with distinct abnormalities \cite{zhang2023challenges}.

Recent advances in foundation models have shown promising potential in addressing these challenges by adopting pathology-agnostic and modality-agnostic strategies. Foundation models, trained on large and diverse datasets, are designed to generalize across a wide range of tasks and imaging conditions, making them highly effective in real-world clinical environments where input heterogeneity is inevitable \cite{lee2024foundation, bao2024foundation}. MedSAM2, a vision transformer-based model, has demonstrated robust performance across multiple anatomical structures and imaging tasks with minimal fine-tuning, thus reducing dependency on task-specific pipelines \cite{zhu2024medicalsam2segment}. Other models such as UNETR, VISION-MAE, and SAM-Med have further explored the utility of large-scale pretraining for enhancing generalizability and segmentation performance across diverse clinical scenarios \cite{otaghsara2025multi, bao2024foundation,liu2024vision}. However, their performance may still require downstream adaptation to align with specific pathology characteristics or domain shifts. These challenges underscore the need for models that not only possess strong generalization capabilities but are also explicitly optimized to segment multiple pathologies in a unified, robust manner \cite{lee2024foundation,otaghsara2025multi,bao2024foundation}.

To address these limitations, by employing a strategy of zero-filled synthetic modalities during training with further training specifications to handle synthesised images, we propose F3-Net, a flexible foundation model capable of handling incomplete modality inputs while segmenting a variety of brain pathologies in a single pass. We discuss the architectural design and training strategies of F3-Net in the  following sections.  
Finally, we present a comprehensive evaluation of F3-Net on the BraTS 2021 glioma, and BraTS 2024 post-treatment glioma and metastasis, and ISLES 2022 datasets, focusing on brain tumor and ischemic stroke segmentation, respectively. We compare the performance of F3-Net to state-of-the-art (SOTA) models under varying modality configurations and full segmentation of pathology scenarios. Through this analysis, we aim to demonstrate the potential of advanced architectures and training specifications in overcoming current limitations in medical image segmentation, ultimately advancing the state of the art and improving clinical outcomes.

\section{Related Works}
Medical image segmentation has advanced significantly with the introduction of deep learning, particularly through encoder-decoder architectures inspired by U-Net. These models have become foundational for pathology detection across various imaging modalities such as MRI, CT, PET, and ultrasound. Kamnitsas et al. introduced the EMMA (Ensemble of Multiple Models and Architectures) framework, which combined 3D CNNs such as DeepMedic, FCN, and U-Net to improve robustness and reduce sensitivity to hyperparameter settings \cite{kamnitsas2017ensembles}. Another major contribution was nnU-Net, a fully automated and self-configuring segmentation pipeline that adapts its architecture, preprocessing, and training strategy to the specifics of a dataset. Its versatility has led to its widespread use across diverse medical segmentation tasks \cite{isensee2021nnu}.

Transformer-based architectures have further advanced the field. UNETR was proposed as a vision transformer (ViT)-based encoder model that processes 3D volumetric patches and decodes them using CNN-based blocks \cite{hatamizadeh2022unetr}. Swin UNETR extended this by replacing standard convolutions with Swin Transformer blocks, enabling efficient high-resolution image processing and better long-range context modeling \cite{hatamizadeh2022swinunetrswintransformers}.

MedNeXt is a more recent architecture that introduces scalable convolutional designs inspired by ConvNeXt. It employs residual up/down-sampling operations and depth-wise convolutions, enabling improved learning efficiency and strong generalization on medical datasets. MedNeXt avoids transformer-based tokenization, instead optimizing convolutional kernels using a progressive depth expansion strategy \cite{roy2023mednext}. The model has shown competitive results across CT and MRI segmentation tasks and continues the trend of combining scalability with architectural efficiency.

The use of ensemble models has become increasingly prominent in medical image segmentation due to their ability to combine the strengths of multiple architectures and mitigate overfitting. For example, recent segmentation pipelines have combined models such as Swin UNETR, nnU-Net, and MedNeXt, leveraging both convolutional and transformer-based representations. Notably, top-performing methods in the BraTS 2023 and 2024 benchmarks adopted this ensemble strategy to achieve superior robustness and accuracy in brain tumor segmentation \cite{ roy2023mednext, brats2023_2024}.

Multimodal and language-guided segmentation models have also seen growing interest. Contrastive pretraining approaches such as MI-Zero \cite{lu2023visuallanguagepretrainedmultiple} and BioViL-T \cite{bannur2023learning} use paired image-text data to enable zero-shot segmentation, enabling models to generalize across different tasks without additional fine-tuning. These approaches have proven useful in histopathology and radiology, although they typically rely on high-quality textual annotations during training.

Following the success of the Segment Anything Model (SAM) in natural image segmentation \cite{kirillov2023segment}, several adaptations have been proposed for medical domains. MedSAM \cite{ma2024segment}, MedLSAM \cite{Lei2023medlam}, and SAM-Med2D \cite{zhang2023sammed2d,wang2025sam} modify SAM's architecture or training objectives to align better with medical data characteristics, such as high-resolution 3D volumes, complex tissue structures, and subtle contrast variations. These adaptations have demonstrated encouraging results in few-shot and prompt-based segmentation of radiology and pathology images.

Despite this progress, many models remain tailored to specific modalities and require task-specific training or fine-tuning. To address this, modality-agnostic and cross-domain learning strategies have been proposed. These include shared latent spaces \cite{karimi2022modality}, contrastive multimodal representation learning \cite{chen2023generalizable}, and unified cross-modal pretraining \cite{li2023unified}, all aiming to enhance generalizability across imaging types.

\section{Dataset}

\subsection{BraTS 2021 Dataset}

The BraTS 2021 dataset is a curated, multi-institutional collection of multi-parametric MRI scans, designed to support both brain tumor segmentation and radiogenomic analysis in pre-operative glioma cases \cite{baid2021rsna}. It includes more than 2000 MRI scans acquired from over 20 institutions, with variations in imaging equipment and acquisition protocols that enhance its representativeness across clinical environments. All cases include histopathologically confirmed glioma diagnoses, and a subset includes MGMT promoter methylation status to support integrated imaging-genomics research.  
Each subject is represented by four MRI modalities: T1, contrast-enhanced T1 (T1-Gd), T2, and T2-FLAIR. Preprocessing steps involved conversion from DICOM to NIfTI \cite{li2016first}, skull stripping, co-registration to a common anatomical space, resampling to 1 mm isotropic resolution, and modality-wise intensity normalization. Tumor segmentation masks were generated using ensemble methods—such as nnU-Net \cite{isensee2021nnu}, DeepScan \cite{mckinley2018ensembles}, and DeepMedic \cite{kamnitsas2015multi,kamnitsas2017efficient}—fused via the STAPLE algorithm \cite{warfield2004simultaneous}, and subsequently refined and validated by experienced neuroradiologists according to a standardized annotation protocol. Annotations include three major tumor sub-regions: enhancing tumor (ET), non-enhancing tumor core (NETC), and peritumoral edema or non-enhancing FLAIR hyperintensity (ED/SNFH), in alignment with VASARI feature definitions.

\subsection{BraTS 2024 Dataset}

The BraTS 2024 dataset is a large-scale, multi-institutional benchmark curated for the evaluation of automated brain tumor segmentation models in real-world clinical settings. It comprises over 2,200 post-treatment glioma cases, 651 pre-treatment brain metastasis cases, and 400 meningioma radiotherapy planning cases, collected from multiple academic medical centers worldwide \cite{deverdier20242024braintumorsegmentation}. Each case includes four standardized MRI modalities—T1-weighted (T1), contrast-enhanced T1-weighted (T1-Gd), T2-weighted (T2), and T2-FLAIR—offering complementary views of neuroanatomy and pathology.  
The glioma cohort captures the complexity of longitudinal post-surgical and post-chemoradiation imaging, with annotations for ET, NETC, SNFH, and resection cavity (RC).  
The metastasis cohort, consisting of secondary brain tumors originating from various primary cancers, is labeled for enhancing tumor, necrotic core and peritumoral edema.  
All MRI scans were originally stored in DICOM format and converted to NIfTI using the dcm2niix tool \cite{li2016first}, preserving metadata and spatial integrity. Preprocessing included skull stripping using the HD-BET algorithm \cite{Isensee_2019}, affine registration to the MNI152 template using the Greedy algorithm via the CaPTk toolkit, resampling to an isotropic voxel size of 1×1×1 mm³, and intensity normalization to reduce cross-scanner variation.  
Segmentation masks were initially produced using an ensemble of five deep learning models, including nnU-Net and SegResNet architectures, and were fused using the STAPLE algorithm \cite{deverdier20242024braintumorsegmentation}. These consensus maps were then reviewed and refined by expert neuroradiologists following a standardized annotation protocol.

\subsection{ISLES 2022 Dataset}

The Ischemic Stroke Lesion Segmentation (ISLES) 2022 dataset is a multi-center, expert-annotated resource designed to advance the development and evaluation of automated segmentation algorithms for acute and subacute ischemic stroke lesions. This dataset comprises a total of 400 magnetic resonance imaging (MRI) cases, collected retrospectively from three distinct stroke centers utilizing various MRI scanners, including 3T Philips (Achieva, Ingenia), 3T Siemens (Verio), and 1.5T Siemens MAGNETOM (Avanto, Aera) systems \cite{hernandez2022isles}.  
Each case in the dataset includes three MRI sequences: Fluid-Attenuated Inversion Recovery (FLAIR), Diffusion-Weighted Imaging (DWI) with a b-value of 1000 s/mm², and the corresponding Apparent Diffusion Coefficient (ADC) map. These sequences are provided in NIfTI format, adhering to the Brain Imaging Data Structure (BIDS) convention, and are released in their native space without prior registration. To ensure patient anonymity, skull-stripping was performed on all images prior to release. Additionally, available scanner metadata extracted from the DICOM headers is provided in JSON format. FLAIR to DWI rigid registration was performed using FSL package, and subsequent skull-stripping of DWI and ADC using the registered brain mask was performed. 

Expert neuroradiologists manually annotated the stroke lesions in each case, providing high-quality ground truth labels for supervised learning. The dataset encompasses a wide range of lesion sizes, quantities, and locations, reflecting the heterogeneity observed in clinical practice.  
The ISLES 2022 dataset serves as a benchmark for the development of reliable and accurate ischemic stroke lesion segmentation methods.

\subsection{White Matter Hyperintensity Dataset}
The White Matter Hyperintensity (WMH) Lesion Dataset offers 3D T1-weighted and 2D FLAIR MRI scans from multiple clinical institutions, reflecting variability in scanners, protocols, and manual annotations. Preprocessing steps include skull stripping, bias correction, and resampling to 1 mm isotropic resolution, with FLAIR scans standardized to 3 mm slice thickness. The dataset comprises 170 cases acquired on five different scanners from three vendors.

Lesions were manually segmented following STandards for ReportIng Vascular changes on nEuroimaging (STRIVE) criteria. An expert observer performed initial delineations using contour tracing, followed by peer review and correction by a second expert. These reviewed and corrected masks serve as the reference standard.

The final segmentations were converted into binary masks, where voxels with $>$50\%   lesion volume were included. WMH regions were assigned label 1, background as label 0, and other pathologies (if present) as label 2—expanded slightly using in-plane dilation. In case of overlap, WMH labels took precedence. Additional inter-observer validation was conducted by two more experts, ensuring annotation reliability \cite{AECRSD2022}.

\subsection{Multiple Sclerosis Lesion Dataset}

The Multiple Sclerosis (MS) Lesion Dataset from the Shifts 2.0 benchmark supports white matter lesion (WML) segmentation using 3D T1-weighted and FLAIR MRI scans. Data were collected from multiple clinical sites using both 1.5T and 3T scanners across various vendors, introducing natural distributional shifts in imaging characteristics.

The dataset combines cases from the MSSEG challenge (MICCAI 2016/2021) and additional real-world scans, providing a diverse and representative sample. Preprocessing includes denoising, skull stripping, bias correction, T1-to-FLAIR registration, and resampling to 1\,mm isotropic resolution. Expert-annotated lesion masks were fused using consensus labeling and interpolated to match the processed image resolution.

Designed to benchmark both in-distribution and out-of-distribution generalization, the dataset enables robust training and evaluation of uncertainty-aware models in clinically variable settings \cite{malinin2022shifts}.

\subsection{Data Pre-Processing}
\begin{figure}[H]  
    \centering
    \includegraphics[width=0.9\textwidth]{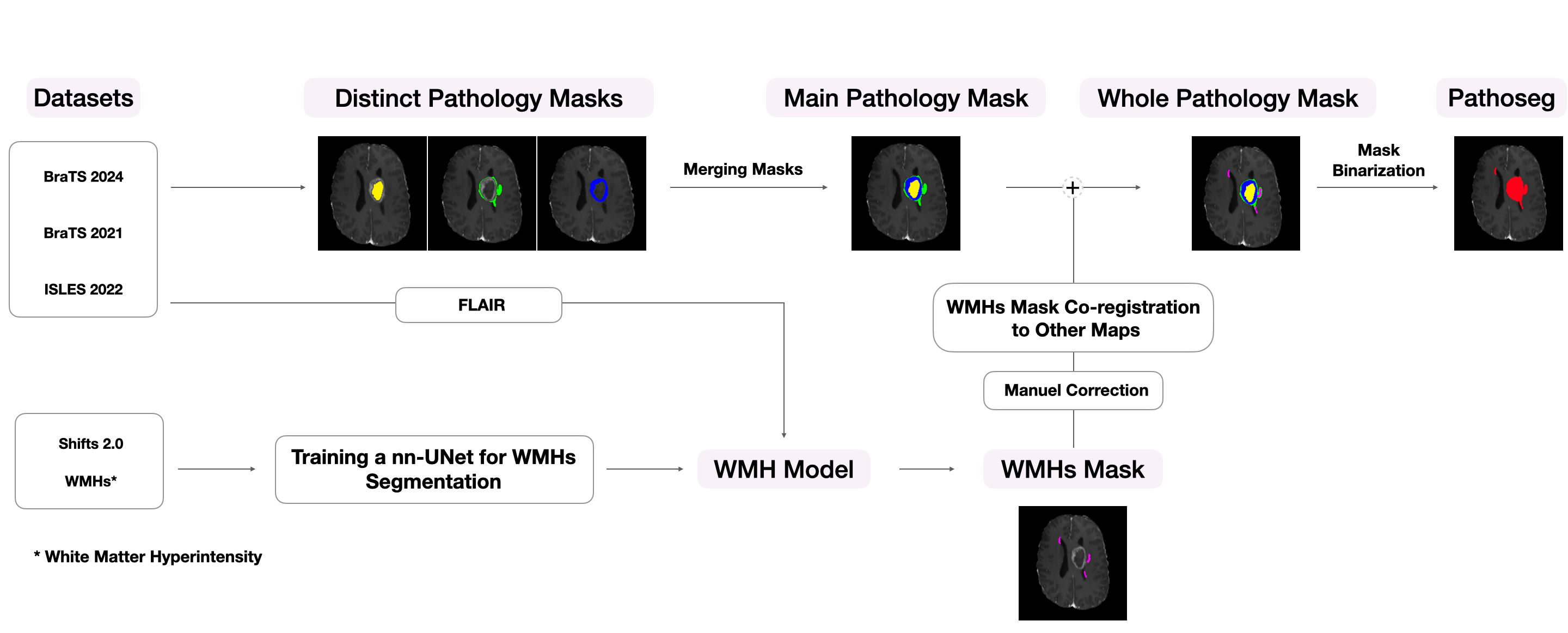}
    \caption {\textbf{ Pipeline for generating comprehensive pathology segmentation masks (Pathoseg).}  This figure depicts the pipeline for generating a unified pathology segmentation mask from multiple neuroimaging datasets. First, a vanilla nnU-Net was trained on WMH data to segment WMHs in FLAIR images from BraTS 2021, BraTS 2024, and ISLES 2022. The generated WMH masks were then manually refined by a neuroradiologist and co-registered to other MRI modalities. The distinct pathology masks already existing in the datasets were merged together. Finally, the co-registered WMH masks and main pathology mask were combined and binarized to produce a comprehensive “Pathoseg” mask that captures all relevant brain pathologies.}
\label{figure-1}

\end{figure}

We trained a vanilla nnU-Net with WMH datasets to automatically segment WMH in FLAIR modality in BraTS 2021, BraTS 2024 and ISLES 2022 datasets. (i) WMH segmentation as model outputs were further manually corrected by a neuroradiologist (RR), (ii) WMH mask on FLAIR space were co-registered to other existing modalities (hereinafter called “co-registered WMH mask”), (iii) the “distinct pathology masks” already existing in datasets were merged together (hereinafter called “main pathology mask”), (iv) the co-registered WMH mask was merged with “main pathology mask” and generates “whole pathology masks”. Then, it will be binarized to generate “Pathoseg” which captures all relevant brain pathologies (Figure \ref{figure-1}). For all the missing modalities among T1, T1-Gd, T2, FLAIR, DWI, and ADC, a zero-filled image (hereinafter referred to as a zero-image) with the same shape as the existing images was synthesized. In Section 3.7, we explain in detail how the model handles zero-images.

\vspace{1mm}

\subsection{Handling Missing Modalities with Zero-Image Synthesis}
To ensure consistent input dimensionality across all subjects, our F3-Net model expects six MRI modalities—T1, T1-Gd, T2, FLAIR, DWI, and ADC—at training and inference time. However, clinical datasets frequently contain missing modalities due to variations in acquisition protocols, scanner availability, or patient-related constraints. To handle such cases, we employ a straightforward zero-image strategy: for each missing modality, a placeholder volume filled entirely with zero intensity values is synthesized. These zero-images are generated with identical spatial dimensions to the available modalities, preserving voxel-wise alignment and enabling seamless integration into the model’s input pipeline. This approach has been employed in prior studies as a baseline for handling missing inputs, where entire channels are set to zero to maintain architectural consistency \cite{shen2019brain,li2014deep}. Additionally, similar concepts are found in early CNN-based segmentation methods using dropout strategies that implicitly zero out modality channels during training \cite{havaei2017brain}.

While this zero-filled approach does not attempt to recover underlying anatomy, it offers a simple and effective mechanism to maintain consistent input structure. In Section~4, we further analyze how the network processes zero-filled inputs and how downstream encoder features are modulated during training.

More recent works have explored alternative strategies for handling missing modalities through data-driven synthesis. Techniques such as masked autoencoding \cite{liu2023m3ae}, GAN-based reconstruction \cite{zhang2023unified, baldini2023pix2pix}, and diffusion-based inpainting \cite{meng2024m2dn} aim to estimate realistic surrogate images for the missing channels. While these approaches offer improved semantic fidelity, they also introduce computational and architectural complexity. Furthermore, these approaches commonly cannot generate modality-specific biological signals critical to pathology detection.
In contrast, our use of zero-images provides a baseline strategy that avoids imputation noise while ensuring robustness during both training and inference.

\section{F3-Net Architecture, Configuration and Implementation }
\begin{figure}[H]  
    \centering
    \includegraphics[width=0.9\textwidth]{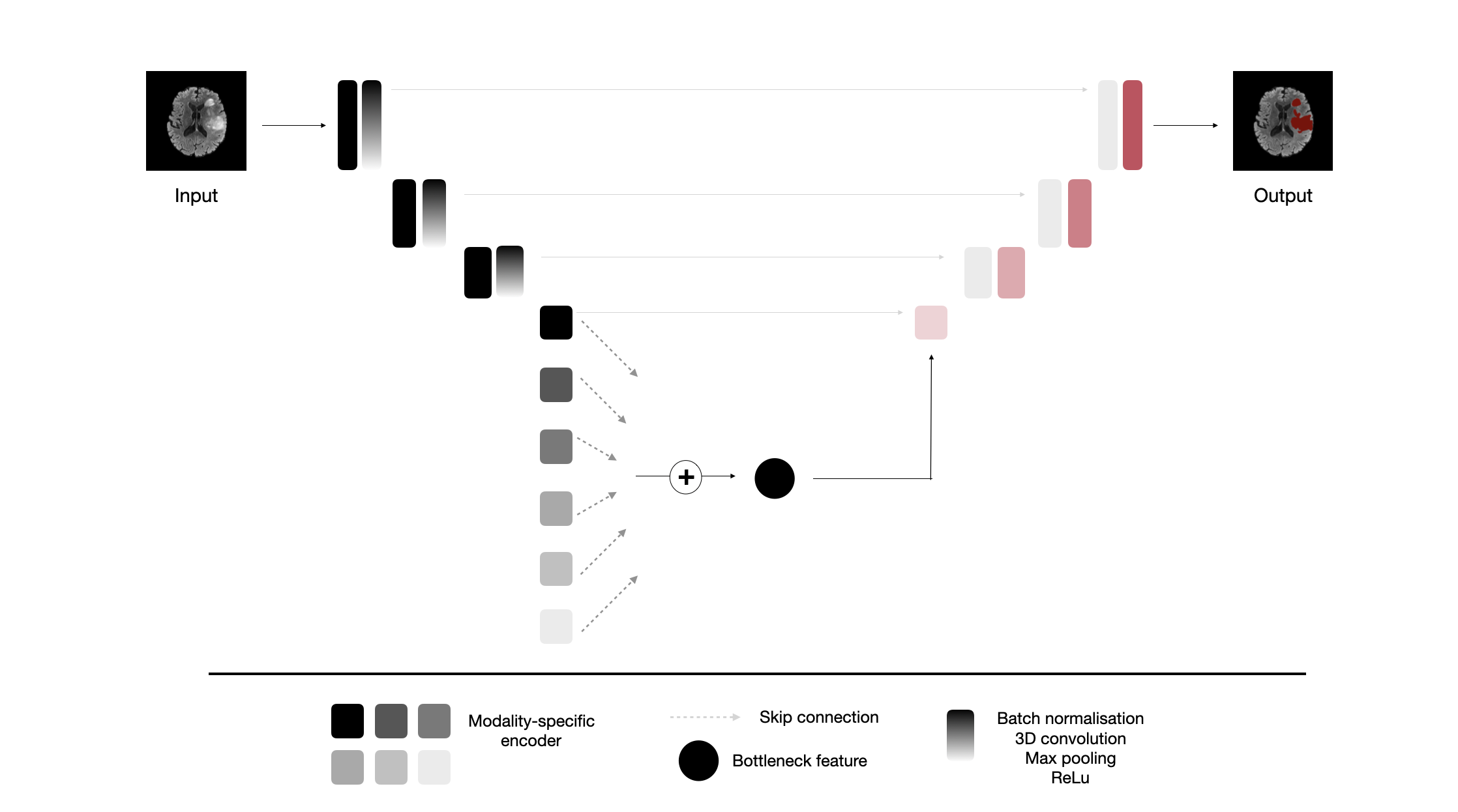}
    \caption {\textbf{ Overview of the F3-Net architecture.}  Each MRI modality is processed by its own dedicated encoder, enabling the extraction of features that are specifically adapted to the distinctive properties of each modality. At the bottleneck stage, the encoded features from all modalities are merged, blending their diverse information to create a unified and robust feature representation. This integrated representation is subsequently processed by a shared decoder, which produces a lesion mask that accurately identifies and outlines all pathological regions across the input MRI modalities.}
\label{figure-2}

\end{figure}

Our approach centers around the nnU-Net framework \cite{isensee2021nnu}, which serves as the foundational architecture for medical image segmentation. In our model, we implement separate encoders tailored for different imaging modalities, while a common decoder is utilized across the board (Figure \ref{figure-2}). Each input image is directed through its respective modality-specific encoder, and then the unified decoder produces anomaly segmentations. Beyond simply filling in missing inputs, we implement a modality-aware mechanism in the nnU-Net architecture to neutralize the effect of zero-images during training. Specifically, for each input case, the feature representations derived from zero-image modalities are forcibly set to zero at the most downstream stage of the encoder (i.e., the deepest encoder layer before the bottleneck). This ensures that these synthetic inputs do not contribute to the learned representations or affect backpropagation. By explicitly masking their influence, the network learns to rely solely on the available informative modalities. This strategy enables robust segmentation performance despite incomplete modality coverage and avoids the potential introduction of artifacts or misleading patterns from synthetic zero inputs. The segmentation task employs the following loss function consisting of two components:

\begin{equation}
\mathcal{L} = \lambda_1 \cdot \mathcal{L}_{\text{Dice}}(s, \hat{s}) + \lambda_2 \cdot \mathcal{L}_{\text{CE}}(s, \hat{s})
\end{equation}

where:
\begin{itemize}
  \item $\mathcal{L}_{\text{Dice}}(s, \hat{s})$ is the Dice loss, which maximizes the overlap between the predicted segmentation $\hat{s}$ and the ground truth label $s$. This term is weighted by the coefficient $\lambda_1$.
  \item $\mathcal{L}_{\text{CE}}(s, \hat{s})$ is the cross-entropy loss, penalizing incorrect pixel-wise predictions to better align the predicted and actual segmentation maps. This is weighted by $\lambda_2$.
\end{itemize}

\subsection*{Training and Implementation Details}

The model is trained with a global batch size of 2 and input patches sized of (80, 112, 80). Optimization is performed using stochastic gradient descent (SGD) with momentum set to 0.95 and weight decay of 3e-5. The initial learning rate at 0.01 with the polynomial decay scheduler with a power of 0.9. Training is conducted for a maximum of 1K epochs.

The model architecture, data preprocessing, and augmentation strategies follow the nnU-Net framework~\cite{isensee2021nnu}.

\section{Results}

\subsection{BraTS Dataset Performance}
To comprehensively evaluate F3-Net’s generalization ability, we conducted inference on two variants of the BraTS dataset: (i) a whole pathology experiment, which includes both the main tumor pathology (gliomas or metastases) and WMHs and (ii) a main pathology experiment containing only the primary disease labels. The whole pathology setup enables direct comparison with "Pathoseg", a gold-standard benchmark for multi-pathology segmentation. In contrast, the main pathology experiment was used to compare F3-Net against other state-of-the-art models that are specifically trained for single-pathology segmentation tasks.

The results show that F3-Net performs strongly across both settings. On the whole pathology dataset, F3-Net achieves average Dice Similarity Coefficients (DSCs) of 94.31\% for BraTS-GLI 2024, 82.07\% for BraTS-MET 2024, and 94.12\% for BraTS 2021 (Table \ref{tab:T1}). On the main pathology dataset, used for fair comparison with Vanilla nnU-Net and Swin UNETR, F3-Net also demonstrates superior performance, as shown in Table \ref{tab:T2}. Notably, F3-Net achieves average DSCs of 93.13\% for BraTS-GLI 2024, 80.95\% for BraTS-MET 2024, and 93.42\% for BraTS 2021 for the main pathology segmentation. These results confirm F3-Net’s adaptability and accuracy in both specialized and broad clinical segmentation scenarios.

\subsection{ISLES Dataset Performance}
In the ISLES 2022 dataset, we evaluated F3-Net under two settings: (i) the whole pathology experiment, which includes both infarct lesions and WMHs and (ii) the main pathology experiment, which focuses solely on infarct lesions. In the whole pathology setting, F3-Net achieves a solid average DSC of 79.92\% (Table \ref{tab:T1}), demonstrating its capacity to segment multiple lesion types simultaneously. Although its performance in the main pathology setting is slightly lower than that of nnU-Net and SegResNet, as shown in Table \ref{tab:isles_comparison}, F3-Net remains competitive—especially given the inherent difficulty of infarct lesion segmentation and the added generalization demands placed on foundation models.

\vspace{0.5cm}
\begin{table}[htbp]
\centering
\small
\begin{tabular}{lccccc}
\toprule
\textbf{Dataset} & \textbf{Av. DSC (\%)} & \textbf{Accuracy} & 
\textbf{Sensitivity} & \textbf{Specificity} & \textbf{Precision} \\

\midrule

BraTS 2024       & 93.95 & 99.88 & 93.50 & 99.94 & 94.87 \\
BraTS-GLI 2024   & 94.31 & 99.89 & 93.97 & 99.95 & 95.22 \\
BraTS-MET 2024   & 82.07 & 99.84 & 80.74 & 99.93 & 86.27 \\
BraTS 2021       & 94.12 & 99.88 & 93.50 & 99.95 & 94.54 \\
ISLES 2022       & 79.92 & 99.66 & 78.22 & 99.84 & 83.81 \\
\bottomrule
\end{tabular}
\caption{Performance of F3-Net on various datasets for the whole pathology segmentation, which includes main tumor pathologies (gliomas, metastases) as well as stroke lesions (ISLES) and WMHs.}
\label{tab:T1}
\end{table}

\vspace{0.5cm}
\begin{table}[htbp]
\centering
\small
\captionsetup{justification=centering}
\begin{tabular}{lccc}
\toprule
\textbf{Methods} & \textbf{BraTS-GLI 2024} & \textbf{BraTS-MET 2024} & \textbf{BraTS 2021} \\
\midrule
F3-Net Av. DSC (\%)           & 93.13 & 80.95 & 93.42 \\
Vanilla nnU-Net Av. DSC (\%)  & 88.91 & 72.82 & 90.89 \\
Swin UNETR Av. DSC (\%)       & 82.04 & 61.17 & 91.80 \\
MedNeXt Av. DSC (\%)          & 86.14 & 70.66 & 90.55 \\
\bottomrule
\end{tabular}
\caption{ Comparison of DSC (\%) of different methods on BraTS-GLI 2024, BraTS-MET 2024, and BraTS 2021 datasets, using only the main pathology labels (gliomas or metastases), excluding additional lesions such as WMHs.} \vspace{4mm}
\label{tab:T2}
\end{table}

\begin{table}[htbp]
  \centering
  \begin{tabular}{l c}
    \toprule
    \textbf{Methods} & \textbf{Av. DSC (\%)} \\
    \midrule
    F3-Net     & 77.28 \\
    nnU-Net    & 81.87 \\
    SegResNet  & 82.23 \\
    \bottomrule
  \end{tabular}
  \caption{Performance comparison of different methods on the ISLES 2022 dataset using only the main pathology (infarct lesions), excluding additional lesions such as WMHs.}
  \label{tab:isles_comparison}
\end{table}

\vspace{5mm}

\section{Discussion}
F3-Net addresses several critical limitations that have long challenged the clinical adoption of deep learning-based medical image segmentation: (i) dependence on complete multimodal inputs, (ii) poor generalizability across diverse clinical settings, and (iii) narrow task specificity.
The robust performance of F3-Net across heterogeneous datasets—including BraTS 2021, BraTS 2024 (glioma, metastasis), and ISLES 2022—suggests that foundation models, when carefully optimized, can meaningfully bridge the translational gap between algorithmic development and clinical deployment.

\textbf{Flexibility through Synthetic Training} 
A key innovation in F3-Net lies in its flexible handling of incomplete input modalities. In routine clinical workflows, acquisition of all six standard MRI sequences (T1, T1-Gd, T2, FLAIR, DWI and ADC) may be impeded by patient motion, resource constraints, or scanning time limitations \cite{chen2023deep, stucht2015highest}. Traditional models either fail outright or experience significant performance drops in such cases \cite{jiang2022swinbts, yan2022seresu}.
F3-Net circumvents this limitation by leveraging zero-filled synthetic modality during training and inference, allowing the model to learn robust cross-modal representations even in the absence of specific channels. This approach mirrors strategies seen in generative modality imputation frameworks \cite{chartsias2019factorised, zheng2024selfsupervised3dpatientmodeling, billot2023synthseg}, but uniquely avoids the dependency on explicit synthesis networks (e.g., GANs), which often hallucinate unrealistic features and degrade diagnostic reliability \cite{shin2021deep, shin2018medical}.

\textbf{Generalization Across Pathologies}
Another noteworthy contribution of F3-Net is its demonstrated ability to segment a wide spectrum of brain abnormalities—ranging from gliomas and metastases to ischemic stroke lesions and WMHs— without retraining for each condition. Most segmentation models are trained in a disease-specific manner, limiting their utility when deployed in settings where a broad differential diagnosis is required \cite{kamnitsas2017efficient}. Even advanced vision transformer models like UNETR and Swin UNETR often require task-specific fine-tuning or ensembling to maintain competitive performance \cite{hatamizadeh2022unetr, hatamizadeh2022swinunetrswintransformers}. In contrast, F3-Net’s unified training on multiple pathologies echoes the objectives of recent generalist models like MedSAM2 \cite{zhu2024medicalsam2segment} and VISION-MAE \cite{liu2024vision}, yet it goes further by integrating architectural and training-level adaptations to support direct multi-pathology segmentation in a single forward pass.

\textbf{Performance Under Domain Shift and Data Heterogeneity}
The robustness of F3-Net across clinical variability is partly attributable to the inclusion of diverse datasets during training and validation. The BraTS 2024 and ISLES 2022 datasets, for example, are sourced from multiple institutions with differing scanners, field strengths, and imaging protocols, reflecting real-world conditions \cite{deverdier20242024braintumorsegmentation,hernandez2022isles, baid2021rsnaasnrmiccaibrats2021benchmark}. These inherent domain shifts are known to degrade model performance in conventional pipelines \cite{guan2021domain}.
F3-Net’s resilience in this context is evidence of its successful generalization capacity, aligning with previous observations in the Shifts 2.0 dataset and associated studies on distributional robustness in medical imaging \cite{malinin2022shifts}.
This all-in-one model approach enhances the efficiency and versatility of the segmentation process, providing a comprehensive solution for clinical use.

\bibliographystyle{IEEEtran}  
\bibliography{references}  

\end{document}